\documentclass[12pt]{article}
\pdfoutput=1 
\usepackage{graphicx} 
\PassOptionsToPackage{square,numbers,sort&compress}{natbib}
\usepackage[preprint]{neurips_2024}
\usepackage[utf8]{inputenc} 
\usepackage[T1]{fontenc}    
\usepackage{hyperref}       
\hypersetup{
    colorlinks=true,        
    linkcolor=black,        
    citecolor=blue,         
    urlcolor=blue,          
    filecolor=blue          
}
\usepackage{url}            
\usepackage{newtxtext,newtxmath} 
\usepackage{booktabs}       
\usepackage{amsfonts}       
\usepackage{nicefrac}       
\usepackage{svg}            
\usepackage{microtype}      
\usepackage{multirow}       
\usepackage[table,x11names]{xcolor}
\usepackage{multirow}       
\usepackage{float}          
\usepackage{appendix}       
\usepackage{setspace}       
\usepackage[hang,flushmargin]{footmisc}
\usepackage{pdfpages}
\usepackage{xcolor}
\definecolor{lightergray}{gray}{0.9}

\usepackage{array}

\usepackage{pgfplots}
\pgfplotsset{width=10cm,compat=1.9}

\usepackage{tikz}

\begin{document}

\title{Litespark Technical Report: \\ High-Throughput, Energy-Efficient LLM Training Framework}
\author{Nii Osae Osae Dade
$\quad$ Moinul Hossain Rahat\\
\\
Mindbeam AI \thanks{Correspondence to: {research@mindbeam.ai}} \\
}

\maketitle

\begin{abstract}
    Training Large Language Models (LLMs) is plagued by long training times and massive energy consumption, with modern models requiring months of computation and gigawatt-hours of electricity. In light of these challenges,we introduce Litespark, a novel pre-training framework that addresses these inefficiencies through targeted optimizations to transformer attention and MLP layers. Our approach combines architectural improvements with algorithmic enhancements to maximize Model FLOPs Utilization (MFU) while maintaining compatibility with standard transformer implementations. Comprehensive benchmarking on 3B and 30B parameter Llama models using the SlimPajama-627B dataset demonstrates substantial performance gains: 2x--6x training throughput improvement and $55\%-83$\% energy consumption reduction across multi-node H200 GPU clusters. These optimizations are model- and hardware-agnostic, enabling broad applicability across transformer architectures and extending to post-training phases including supervised fine-tuning and direct preference optimization.  
    
\end{abstract}

\section{Introduction}
The exponential growth of large language models (LLMs) since GPT-3's release in 2020 \cite{DBLP:conf/nips/BrownMRSKDNSSAA20} has fundamentally transformed the way we use artificial intelligence (AI) in daily lives. Since then, a plethora of LLMs have been developed \cite{DBLP:journals/corr/abs-2211-05100, DBLP:journals/corr/abs-2302-13971, 
claude, 
DBLP:journals/corr/abs-2303-08774,  DBLP:journals/corr/abs-2305-10403, 
DBLP:journals/corr/abs-2307-09288, 
claude2, 
DBLP:journals/corr/abs-2309-16609, 
DBLP:journals/corr/abs-2310-06825, DBLP:journals/corr/abs-2311-16867,
DBLP:journals/corr/abs-2312-11805,
DBLP:journals/corr/abs-2412-19437,
DBLP:journals/corr/abs-2501-12948, claude3.7, o3,  meta2025llama4, grok3, claude4.5}, demonstrating significant progress towards the pursuit of artificial general intelligence (AGI). However, this progress has come at an unprecedented computational and environmental cost. Training modern LLMs now requires months of compute time, tens of millions of dollars, and energy consumption equivalent to powering thousands of homes for years \cite{DBLP:journals/jmlr/LuccioniVL23}. 

The scale of this challenge has intensified over the years. For example, GPT-3 175B model was trained for 3,640 PF-days \cite{DBLP:conf/nips/BrownMRSKDNSSAA20}, which would take around $50-70$ days on $1,000$ NVIDIA V100 GPUs. Llama 3.1-405B reportedly consumed $30.84$ million GPU-hours \cite{nvidia2024Llama31modelcard}, equivalent to $80$ days of training on $16,000$ H100 GPUs. Simultaneously, energy consumption has exploded: from GPT-3's $1,287$ MWh \cite{DBLP:journals/corr/abs-2104-10350} to Llama 3.1-405B's approximately 21.6 GWh (based on 700 W consumption per H100 GPU) \cite{nvidia2024Llama31modelcard}. The environmental impact of model training has increased dramatically. Llama-3.1-405B's training would leave a carbon footprint reaching 8,930 tonnes CO2eq \cite{nvidia2024Llama31modelcard}, representing a 16-fold increase over GPT-3 175B's  552.1 tonnes CO2eq \cite{DBLP:journals/corr/abs-2104-10350}. 

Both of these challenges -- extended training times and massive energy consumption -- stem from the same fundamental issue: inefficient utilization of computational resources during transformer training. Despite consuming full power, GPUs during standard LLM pre-training often operate at suboptimal utilization rates of $30\%-50\%$. This inefficiency creates a compound problem: training takes longer than necessary while simultaneously wasting energy: organizations face both extended time-to-market delays and inflated energy costs. Running ablations in the model development phase becomes slower and prohibitively costly, effectively limiting the breadth of scientific exploration. 

The suboptimal performance in LLM pre-training stems largely from bottlenecks in the core transformer architecture, particularly in the attention and MLP (Multi-Layer Perceptron) layers that constitute the majority of computational operations. The attention mechanism \cite{DBLP:conf/nips/VaswaniSPUJGKP17} suffers from inherent limitations in memory bandwidth that prevent GPUs from achieving maximum computational throughput. Traditional attention implementations are memory-bound rather than compute-bound, causing expensive GPU compute units to remain idle while waiting for data transfers \cite{DBLP:conf/nips/DaoFERR22}. Similarly, standard MLP layers often fail to fully utilize modern GPU capabilities, particularly the specialized Tensor Core units designed for high-throughput operations \cite{nvidia2023tensorcore}. These architectural inefficiencies translate directly into wasted energy: every second a GPU operates below capacity represents energy consumed without proportional computational gains.

Recent research has demonstrated that algorithmic improvements can address both challenges simultaneously. Techniques like FlashAttention achieve $2$x--$3$x training speedup while maximizing GPU utilization \cite{DBLP:conf/nips/DaoFERR22, DBLP:conf/iclr/Dao24, DBLP:conf/nips/ShahBZTRD24}. Mixture-of-Experts approaches reduce training time and computational requirements by 4x--7x through sparse activation patterns \cite{DBLP:journals/jmlr/FedusZS22}. 

In this technical report, we introduce Litespark, a novel pre-training framework that simultaneously addresses both training time and energy efficiency challenges through targeted optimizations to the transformer architecture's attention and MLP layers. Our approach focuses on maximizing Model FLOPs Utilization (MFU) while maintaining compatibility with standard transformer implementations. The optimizations occur in two steps. 

\begin{itemize}
    \item Architectural optimization: optimizes the attention and MLP blocks in the transformer architecture.
    \item Algorithmic optimization: optimizes the forward and backward pass operations to increase FLOPs per GPU.
\end{itemize}

Litespark offers 2x--6x enhancement in training throughput, and $55\%-83$\% reduction in the energy consumption during the pre-training process. Notably, these optimizations add on top of the performance improvements from known existing techniques like flash-attention, quantization, model pruning etc. Furthermore, the optimizations are model- and hardware-agnostic, and can be incorporated into any model architectures and hardware families including GPUs and ASICs.  

The report is organized as follows. In section \ref{sec:2}, we describe the setup for running benchmarking experiments comparing the performance of pre-training models with Litespark vs. Llama baselines. Section \ref{sec:3} showcases the main results in terms of enhanced throughput and energy efficiency. Section \ref{sec:4} points out some future directions of research, and we conclude in section \ref{sec:5}.

\section{Experimental Setup} \label{sec:2}
To evaluate the effectiveness of the Litespark framework, we conducted comprehensive benchmarking experiments comparing our optimized implementation against baseline Llama models across multiple scales and configurations. Our experimental design focuses on measuring both training acceleration and energy efficiency improvements while ensuring fair comparison through identical model architectures, datasets, and training hyperparameters. The evaluation covers scenarios from single-node training to large-scale distributed setups, enabling assessment of how our optimizations perform across the full spectrum of practical deployment scenarios.

\subsection{Hardware infrastructure}
All experiments were conducted on an Amazon SageMaker Hyperpod cluster equipped with NVIDIA H200 GPUs. The H200 represents a recent generation of data center GPUs, featuring 141GB of HBM3e memory and peak theoretical performance of 989 TFLOPS for BF16 operations \cite{nvidia2024h200}. Our multi-node distributed training setup utilized high-bandwidth InfiniBand interconnects for intra-node communication between GPUs and AWS Elastic Fabric Adapter (EFA) with NCCL for inter-node communication to minimize communication overhead during parameter synchronization.

We evaluated scalability across multiple node configurations: 1, 2, 16, 32, and 64 nodes for different model sizes. Each node contained 8 H200 GPUs, enabling evaluation of training performance from single-node (8 GPUs) to large-scale distributed scenarios (512 GPUs for the largest configuration). This range allows us to assess both the baseline efficiency improvements and how our optimizations scale with increasing distributed training complexity.

\subsection{Dataset}
Training was performed on the SlimPajama-627B dataset \cite{cerebras2023slimpajama}, a refined version of the RedPajama dataset \cite{DBLP:conf/nips/WeberFAOAALNYAA24}, containing 627 billion tokens of high-quality text data. SlimPajama consists of web pages, books, academic papers, code repositories, and reference materials, representing a diverse and representative sample for general-purpose language model pre-training. The dataset has been preprocessed to remove low-quality content and deduplicated to improve training efficiency. This dataset choice allows for direct comparison with other published LLM training results while ensuring sufficient scale to evaluate performance across extended training runs.

\subsection{Tokenizer}
The training dataset was pre-processed utilizing  
a SentencePiece-based tokenizer \cite{DBLP:conf/emnlp/KudoR18} with a vocabulary size of 32,000 tokens. This tokenizer choice ensures consistency with the Llama model family and enables direct performance comparisons without introducing tokenization-related variations. The tokenizer employs byte-pair encoding (BPE) \cite{10.5555/177910.177914} to handle out-of-vocabulary words and maintains compatibility with the original Llama tokenization scheme, ensuring that our optimizations can be fairly evaluated against baseline implementations using identical text preprocessing.

\subsection{Model architecture}
We have chosen two model configurations based on the Llama architecture to enable direct performance comparison, as shown in Table~1.

\begin{table}[ht]
  \label{table:model-config}
  \centering
  \begin{tabular}{lccccc}
    \toprule
    \textbf{Model size} & \textbf{n\_layers} & \textbf{hidden\_size} & \textbf{n\_heads} & \textbf{kv\_heads} & \textbf{intermediate\_size}\\
    \midrule
    3B & 28 & 2,048 & 16 & 2 & 11,008\\
    30B & 60 & 6,656 & 64 & 64 & 17,920\\
    \bottomrule \\
  \end{tabular}
    \caption{Model configuration}
\end{table}

Both models utilize standard Llama architectural components including RMSNorm for layer normalization \cite{DBLP:conf/nips/ZhangS19a}, SwiGLU activation functions \cite{DBLP:journals/corr/abs-2002-05202} in the MLP layers, and Rotary Positional Embeddings (RoPE) for position encoding \cite{DBLP:journals/ijon/SuALPBL24}. The 3B model employs grouped query attention (GQA) \cite{DBLP:conf/emnlp/AinslieLJZLS23} to reduce memory overhead, while the 30B model uses standard multi-head attention. These configurations were chosen to represent both smaller models suitable for research experimentation and larger models representative of production deployments.

\subsection{Pre-training configuration}
\subsubsection{Distributed training setup}
We employed a combination of data parallelism and tensor parallelism to distribute training across multiple nodes and GPUs \cite{ultrascale_playbook, liang2025torchtitan}. Data parallelism replicates the model across different GPU groups, while tensor parallelism splits individual layers across GPUs to handle models that exceed single-GPU memory capacity.

\subsubsection{Optimizer settings}
All models were trained in BF16 mixed precision using the AdamW optimizer \cite{DBLP:conf/iclr/LoshchilovH19} with $\beta_1 = 0.90$, $\beta_2 = 0.95$, weight decay of 0.01, and gradient clipping threshold of 1.0. We used ZeRO Stage 1 optimization \cite{DBLP:conf/sc/RajbhandariRRH20} to distribute optimizer states across GPUs while maintaining model replicas.

\subsubsection{Learning rate schedule}
Training employed a cosine learning rate scheduler with maximum learning rate of $1.2 \times 10^{-3}$, minimum learning rate of $1.0 \times 10^{-5}$, and $2,000$ warmup steps. The global batch size was set to $256$ across all configurations, with micro-batch sizes adjusted based on memory constraints and node configuration.

\begin{table}[ht]
  \label{table:pretrain-hyperparameters}
  \centering
  \begin{tabular}{ll}
    \toprule
    \textbf{Hyperparameter} & \textbf{Value}\\
    \midrule
    Optimizer & AdamW ($\beta_1 = 0.90$, $\beta_2 = 0.95$)\\
    ZeRO stage & 1\\
    Learning rate scheduler & Cosine \\
    Max learning rate & $1.2 \times 10^{-3}$ \\
    Min learning rate & $1.0 \times 10^{-5}$ \\
    Warmup steps & $2,000$ \\
    Batch size & $256$ \\
    Weight decay & $0.01$ \\
    Gradient clipping threshold & $1.0$ \\
    \bottomrule \\
  \end{tabular}
    \caption{Pre-training hyperparameters}
\end{table}

\subsubsection{Evaluation metrics}
We measured training throughput (tokens per second), computational efficiency (TFLOPs per GPU), training time per iteration, Model FLOPs Utilization (MFU), and total energy consumption in MWh per 500 billion tokens processed. Energy measurements were calculated by integrating GPU power consumption over training time as reported by Wandb telemetry \cite{wandb}. Values represent direct GPU energy consumption during training.

\section{Results} \label{sec:3}
\subsection{Training throughput acceleration}
Litespark delivers substantial reductions in training time across all configurations, directly addressing the time-to-market challenges facing LLM development. For the 3B parameter model, as shown in Table 3, our framework processes 2x--4x more tokens per second than baseline Llama implementations, translating to proportional reductions in total training time. For example, with 8 H200 GPUs, completing a fixed amount of training that would take Llama 100 hours would require only 50 hours with Litespark.

The time savings become more pronounced at scale, where distributed training traditionally suffers from communication bottlenecks. With 128 GPUs, Litespark's $3.81$x speedup for the 3B model means training jobs that previously required weeks can be completed in days. For the 30B model, as shown in Table 4, the 4.73x--6.36x acceleration transforms month-long training cycles into week-long iterations, fundamentally changing the pace of model development and experimentation.

These training time reductions have immediate strategic value beyond energy considerations. Faster training enables rapid iteration during model development, allowing researchers to test more architectural variations and hyperparameter configurations within fixed time budgets. Organizations can respond more quickly to market demands, reduce time-to-deployment for new models, and maintain competitive advantages through faster innovation cycles. The ability to complete training in days rather than weeks also reduces the risk of infrastructure failures derailing long-running experiments.

\begin{table}[htbp]
\centering
\label{tab:throughput}
\begin{tabular}{c l r rrrr}
\toprule
\multicolumn{1}{c}{\textbf{Num}} & \multicolumn{1}{c}{\textbf{Model}} & \multicolumn{1}{c}{\textbf{tokens/sec}} & \multicolumn{1}{c}{\textbf{TFLOPs/}} & \multicolumn{1}{c}{\textbf{time/}} & \multicolumn{1}{c}{\textbf{MFU}} & \multicolumn{1}{c}{\textbf{Speedup}} \\
\multicolumn{1}{c}{\textbf{GPUs}} & \multicolumn{1}{c}{} & \multicolumn{1}{c}{} & \multicolumn{1}{c}{\textbf{GPU}} & \multicolumn{1}{c}{\textbf{iteration (sec)}} & \multicolumn{1}{c}{\textbf{(\%)}} & \multicolumn{1}{c}{} \\
\midrule
\multirow{2}{*}{8} & Litespark & 439,644.81 & 888.06 & 2.38 & 89.35 & \multirow{2}{*}{2.00} \\
 & Llama & 218,967.60 & 442.30 & 6.44 & 44.70 & \\
\midrule
\multirow{2}{*}{16} & Litespark & 862,899.88 & 871.51 & 1.40 & 88.65 & \multirow{2}{*}{2.17} \\
 & Llama & 396,768.52 & 400.73 & 5.49 & 40.63 & \\
\midrule
\multirow{2}{*}{128} & Litespark & 1,387,342.21 & 175.15 & 2.23 & 17.66 & \multirow{2}{*}{3.81} \\
 & Llama & 364,328.66 & 45.99 & 7.19 & 4.67 & \\
\midrule
\multirow{2}{*}{256} & Litespark & 964,981.24 & 61.58 & 2.43 & 6.29 & \multirow{2}{*}{2.25} \\
 & Llama & 428,056.25 & 27.31 & 7.63 & 2.73 & \\
\bottomrule \\
\end{tabular}
\caption{Pre-training throughput of 3B models on H200s}
\end{table}

\begin{table}[htbp]
\centering
\label{tab:throughput_30b}
\begin{tabular}{c l r rrrr}
\toprule
\multicolumn{1}{c}{\textbf{Num}} & \multicolumn{1}{c}{\textbf{Model}} & \multicolumn{1}{c}{\textbf{tokens/sec}} & \multicolumn{1}{c}{\textbf{TFLOPs/}} & \multicolumn{1}{c}{\textbf{time/}} & \multicolumn{1}{c}{\textbf{MFU}} & \multicolumn{1}{c}{\textbf{Speedup}} \\
\multicolumn{1}{c}{\textbf{GPUs}} & \multicolumn{1}{c}{} & \multicolumn{1}{c}{} & \multicolumn{1}{c}{\textbf{GPU}} & \multicolumn{1}{c}{\textbf{iteration (sec)}} & \multicolumn{1}{c}{\textbf{(\%)}} & \multicolumn{1}{c}{} \\
\midrule
\multirow{2}{*}{256} & Litespark & 471,486.24 & 393.25 & 2.23 & 39.54 & \multirow{2}{*}{4.73} \\
 & Llama & 99,604.57 & 83.08 & 10.53 & 8.43 & \\
\midrule
\multirow{2}{*}{512} & Litespark & 508,260.62 & 215.78 & 2.07 & 21.88 & \multirow{2}{*}{6.36} \\
 & Llama & 79,891.47 & 33.32 & 13.09 & 3.38 & \\
\bottomrule \\
\end{tabular}
\caption{Pre-training throughput of 30B models on H200s}
\end{table}

\begin{figure}[htbp]
\centering
\scalebox{0.90}{%
  \includegraphics[width=\textwidth]{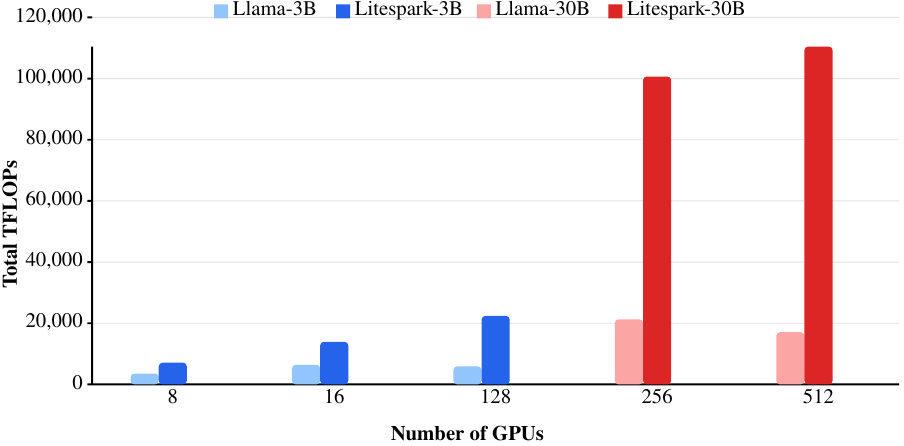}
}
\caption{Pre-training throughput comparison on H200s}
\label{fig:tflops_comparison}
\end{figure}

\subsection{Computational efficiency and resource utilization}
The dramatic training time improvements in Litespark stem from enhanced computational efficiency and resource utilization. This is manifest from the Tera-FLOPs per GPU and Model FLOPs Utilization (MFU) reported in Tables 3 and 4. Litespark achieves 89.35\% MFU compared to Llama's 44.70\% on training with 8 GPUs, indicating our optimizations successfully extract maximum computational value from available hardware. This high utilization rate means that expensive GPU resources operate at near-peak capacity rather than sitting idle due to architectural bottlenecks.

At larger scales, Litespark maintains superior efficiency even as the complexity of distributed training increases. With 128 GPUs, Litespark sustains 17.66\% MFU while Llama drops to 4.67\%, demonstrating that its optimizations address fundamental bottlenecks in multi-node transformer training. For the 30B model at 256 GPUs, Litespark achieves 39.54\% MFU compared to Llama's 8.43\% MFU.

In terms of total computational throughput, as shown in Figure 1, Litespark consistently outperforms the baseline Llama implementation, achieving 2x--6x higher total TFLOPs across all GPU configurations. 
These throughput metrics indicate that our architectural and algorithmic improvements become increasingly valuable for larger models where memory bandwidth limitations traditionally become more severe.

This consistent high utilization across configurations transforms the economics of GPU usage. By achieving $17\%-40$\% MFU compared to Llama's $3\%-8$\% MFU in large-scale configurations, Litespark converts previously wasted computational cycles into productive training progress, maximizing return on infrastructure investment.

\subsection{Energy efficiency}
The throughput improvements directly translate into substantial energy savings. For the 3B model, as shown in Table 5, Litespark reduces energy consumption by $55\%-70$\% across different GPU configurations. Training 500B tokens requires only $0.79-3.41$ MWh with Litespark compared to $1.75-8.01$ MWh with baseline Llama. In particular, energy savings increase with scale: while training with only 8 GPUs shows 55\% energy reduction, training with 128 GPUs achieves 70\% energy savings. Using the standard conversion formula \cite{epa_ghg_calc_2024},
\begin{align*}
\textrm{CO}_2\textrm{eq (tonnes)} &= \textrm{Energy (MWh)} \times \textrm{carbon intensity (kg CO}_2\textrm{eq/kWh)} / 1000
\end{align*}
with an average US carbon intensity of 0.35 kg CO$_2$eq/kWh \cite{epa_egrid_2025}, these energy savings directly translate into carbon emission reductions. For the 3B model, training 500B tokens produces only $0.28-1.19$ tonnes of CO$_2$eq with Litespark versus $0.61-2.80$ tonnes with Llama, as shown in Figure 2 (left).

The 30B model demonstrates even more dramatic gains in energy efficiency, with a $75\%-83$\% reduction in energy, as shown in Table 6. Training 500B tokens on 256 GPUs requires 125.35 MWh with Litespark versus 732.08 MWh with Llama yields an 83\% reduction representing over 600 MWh in savings. This corresponds to 43.87 tonnes of CO$_2$eq with Litespark as compared to 256.23 tonnes with Llama -- a reduction of over 212 tonnes of CO$_2$eq per 500B tokens. At the largest scale (512 GPUs), Litespark consumes $189.47$ MWh compared to Llama's $751.75$ MWh, maintaining 75\% energy savings even with increased communication overhead. The corresponding carbon emissions are 66.31 tonnes versus 263.11 tonnes, as illustrated in Figure 2 (right).

These energy savings have immediate practical implications. For a typical 30B model training run requiring several trillion tokens, our framework could reduce energy consumption from gigawatt-hours to hundreds of megawatt-hours, translating to millions of dollars in electricity cost savings and proportional reductions in carbon emissions. At scale, training a 30B model on 10 trillion tokens would result in approximately 1,478 tonnes of CO$_2$eq with Litespark compared to 5,864 tonnes with Llama --- a reduction of over 4,300 tonnes of CO$_2$eq, equivalent to the annual emissions of nearly 860 passenger vehicles \cite{epa_vehicle_emissions_2025}.

\begin{table}[htbp]
\centering
\label{tab:energy_3b}
\begin{tabular}{c l ccc}
\toprule
\multicolumn{1}{c}{\textbf{Num}} & \multicolumn{1}{c}{\textbf{Model}} & \multicolumn{1}{c}{\textbf{Energy (MWh)/}} & \multicolumn{1}{c}{\textbf{CO2eq (tonnes)/}} & \multicolumn{1}{c}{\textbf{Energy savings}} \\
\multicolumn{1}{c}{\textbf{GPUs}} & \multicolumn{1}{c}{} & \multicolumn{1}{c}{\textbf{500B tokens}} & \multicolumn{1}{c}{\textbf{500B tokens}} & \multicolumn{1}{c}{\textbf{(\%)}} \\
\midrule
\multirow{2}{*}{8} & Litespark & 0.79 & 0.28 & \multirow{2}{*}{54.86} \\
 & Llama & 1.75 & 0.61 & \\
\midrule
\multirow{2}{*}{16} & Litespark & 0.80 & 0.28 & \multirow{2}{*}{55.56} \\
 & Llama & 1.80 & 0.63 & \\
\midrule
\multirow{2}{*}{128} & Litespark & 1.65 & 0.58 & \multirow{2}{*}{69.67} \\
 & Llama & 5.44 & 1.90 & \\
\midrule
\multirow{2}{*}{256} & Litespark & 3.41 & 1.19 & \multirow{2}{*}{57.43} \\
 & Llama & 8.01 & 2.80 & \\
\bottomrule \\
\end{tabular}
\caption{Energy consumption of 3B models on H200s}
\end{table}

\begin{table}[htbp]
\centering
\label{tab:energy_30b}
\begin{tabular}{c l ccc}
\toprule
\multicolumn{1}{c}{\textbf{Num}} & \multicolumn{1}{c}{\textbf{Model}} & \multicolumn{1}{c}{\textbf{Energy (MWh)/}} & \multicolumn{1}{c}{\textbf{CO2eq (tonnes)/}} & \multicolumn{1}{c}{\textbf{Energy savings}} \\
\multicolumn{1}{c}{\textbf{GPUs}} & \multicolumn{1}{c}{} & \multicolumn{1}{c}{\textbf{500B tokens}} & \multicolumn{1}{c}{\textbf{500B tokens}} & \multicolumn{1}{c}{\textbf{(\%)}} \\
\midrule
\multirow{2}{*}{256} & Litespark & 125.35 & 43.87 & \multirow{2}{*}{82.88} \\
 & Llama & 732.08 & 256.23 & \\
\midrule
\multirow{2}{*}{512} & Litespark & 189.47 & 66.31 & \multirow{2}{*}{74.80} \\
 & Llama & 751.75 & 263.11 & \\
\bottomrule \\
\end{tabular}
\caption{Energy consumption of 30B models on H200s}
\end{table}

\begin{figure}[htbp]
\centering
\includegraphics[width=\textwidth]{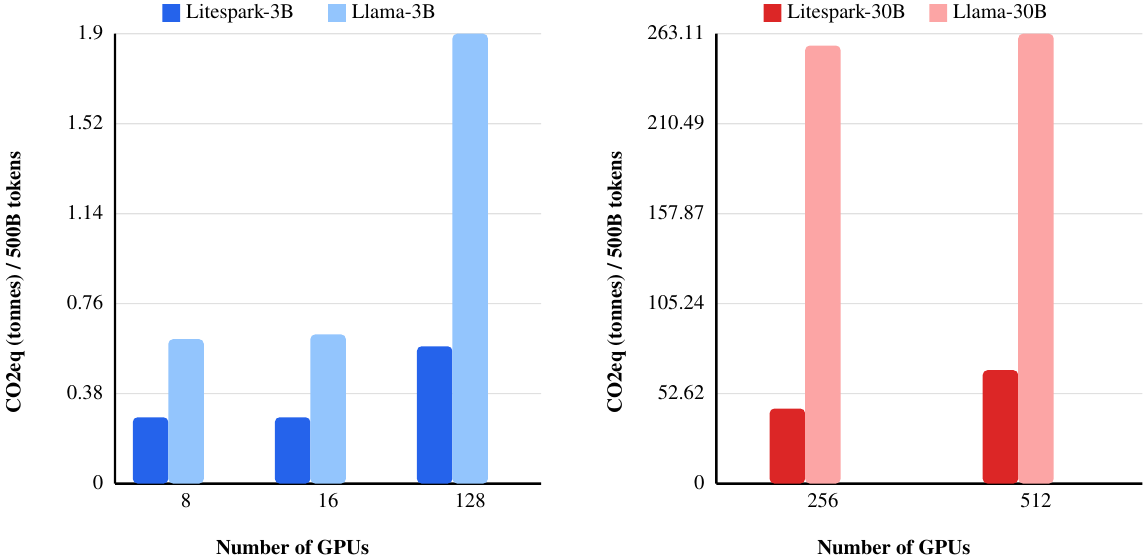}
\caption{$\textrm{CO}_2$ emissions comparison for 3B models (left) and 30B models (right) on H200s}
\label{fig:co2_comparison}
\end{figure}

\section{Future directions} \label{sec:4}
The architectural optimizations demonstrated in the Litespark framework extend far beyond LLM pre-training applications. Here are some of the research directions we are pursing:

\subsection{LLM post-training}
Attention and MLP layer improvements can be applied directly to downstream training phases, including Supervised Fine-Tuning (SFT) and Direct Preference Optimization (DPO) \cite{DBLP:conf/nips/RafailovSMMEF23}, where our preliminary experiments indicate similar performance enhancements to those observed during pre-training. This broad applicability means that efficiency gains compound across the entire model development lifecycle, from initial training through deployment-ready fine-tuning.

\subsection{Foundation models}
Since our optimizations operate at the fundamental transformer block level, they are inherently portable to other transformer-based architectures. The framework can be integrated into multimodal models that utilize encoder-decoder transformers, diffusion models with transformer backbones, and other foundation models based on attention mechanisms. Ongoing experiments show clear promise in throughput enhancement and energy savings in the training of multimodal foundation models. This architectural agnosticism positions Litespark as a foundational optimization that can enhance efficiency across the broader landscape of modern AI systems.

\subsection{Inference}
Early experiments suggest that inference acceleration represents another promising direction. The same architectural improvements that enhance training throughput can potentially reduce inference latency and energy consumption, making deployed models more cost-effective and environmentally sustainable. Given that inference often represents the majority of a model's lifetime energy consumption, these optimizations could have even greater cumulative impact in production environments than during training phases.

\section{Conclusion} \label{sec:5}
We have introduced the Litespark framework demonstrating that targeted architectural optimizations can dramatically reduce both training time and energy consumption in LLM pre-training. Addressing bottlenecks in the attention and MLP layers of the transformer architecture, we have brought down the duration of LLM training by 2--6 times and the energy consumption by $55\%-83$\% compared to the baseline framework. Most importantly, we have demonstrated that faster training and high energy efficiency can be achieved simultaneously without sacrificing model quality or requiring fundamental changes to the model architecture.

The improvement in training throughput represents a paradigm shift for LLM development cycles. This reduces training time from months to days and fundamentally changes the way organizations approach model development, experimentation, and deployment. Litespark's accelerated framework enables rapid iteration, faster response to market needs, and reduced risks associated with long-running computational experiments.

Our findings suggest that the path to sustainable LLM training lies not merely in hardware scaling, but in algorithmic breakthroughs leading to maximal utilization of existing computational resources. Litespark achieves substantial MFU improvements from 3-8\% to 17-40\% in large-scale distributed configurations, heralding a new era of energy-efficient training.

These improvements have implications beyond immediate cost and time savings. Accelerated training democratizes access to large-scale model development by lowering the time barriers that previously constrained experimentation only to institutions with massive computational budgets. At the same time, the $55\%-83$\% energy reductions make previously prohibitive training scenarios economically viable, while addressing environmental sustainability concerns.

Litespark provides a practical pathway toward sustainable and rapid LLM development, where LLM pre-training acceleration and energy efficiency are not just aspirational goals but achieved realities. We are optimistic that these advancements will bring us one step closer to building the next-generation AI infrastructure.


\bibliographystyle{unsrtnat} 
\bibliography{main} 

\begin{thebibliography}{46}
\providecommand{\natexlab}[1]{#1}
\providecommand{\url}[1]{\texttt{#1}}
\expandafter\ifx\csname urlstyle\endcsname\relax
  \providecommand{\doi}[1]{doi: #1}\else
  \providecommand{\doi}{doi: \begingroup \urlstyle{rm}\Url}\fi

\bibitem[Brown et~al.(2020)Brown, Mann, Ryder, Subbiah, Kaplan, Dhariwal, Neelakantan, Shyam, Sastry, Askell, Agarwal, Herbert{-}Voss, Krueger, Henighan, Child, Ramesh, Ziegler, Wu, Winter, Hesse, Chen, Sigler, Litwin, Gray, Chess, Clark, Berner, McCandlish, Radford, Sutskever, and Amodei]{DBLP:conf/nips/BrownMRSKDNSSAA20}
Tom~B. Brown, Benjamin Mann, Nick Ryder, Melanie Subbiah, Jared Kaplan, Prafulla Dhariwal, Arvind Neelakantan, Pranav Shyam, Girish Sastry, Amanda Askell, Sandhini Agarwal, Ariel Herbert{-}Voss, Gretchen Krueger, Tom Henighan, Rewon Child, Aditya Ramesh, Daniel~M. Ziegler, Jeffrey Wu, Clemens Winter, Christopher Hesse, Mark Chen, Eric Sigler, Mateusz Litwin, Scott Gray, Benjamin Chess, Jack Clark, Christopher Berner, Sam McCandlish, Alec Radford, Ilya Sutskever, and Dario Amodei.
\newblock {Language Models are Few-Shot Learners}.
\newblock In Hugo Larochelle, Marc'Aurelio Ranzato, Raia Hadsell, Maria{-}Florina Balcan, and Hsuan{-}Tien Lin, editors, \emph{Advances in Neural Information Processing Systems 33: Annual Conference on Neural Information Processing Systems 2020, NeurIPS 2020, December 6-12, 2020, virtual}, 2020.
\newblock URL \url{https://proceedings.neurips.cc/paper/2020/hash/1457c0d6bfcb4967418bfb8ac142f64a-Abstract.html}.

\bibitem[Scao et~al.(2022)Scao, Fan, Akiki, Pavlick, Ilic, Hesslow, Castagn{\'{e}}, Luccioni, Yvon, Gall{\'{e}}, Tow, Rush, Biderman, Webson, Ammanamanchi, Wang, Sagot, Muennighoff, del Moral, Ruwase, Bawden, Bekman, McMillan{-}Major, Beltagy, Nguyen, Saulnier, Tan, Suarez, Sanh, Lauren{\c{c}}on, Jernite, Launay, Mitchell, Raffel, Gokaslan, Simhi, Soroa, Aji, Alfassy, Rogers, Nitzav, Xu, Mou, Emezue, Klamm, Leong, van Strien, Adelani, and et~al.]{DBLP:journals/corr/abs-2211-05100}
Teven~Le Scao, Angela Fan, Christopher Akiki, Ellie Pavlick, Suzana Ilic, Daniel Hesslow, Roman Castagn{\'{e}}, Alexandra~Sasha Luccioni, Fran{\c{c}}ois Yvon, Matthias Gall{\'{e}}, Jonathan Tow, Alexander~M. Rush, Stella Biderman, Albert Webson, Pawan~Sasanka Ammanamanchi, Thomas Wang, Beno{\^{\i}}t Sagot, Niklas Muennighoff, Albert~Villanova del Moral, Olatunji Ruwase, Rachel Bawden, Stas Bekman, Angelina McMillan{-}Major, Iz~Beltagy, Huu Nguyen, Lucile Saulnier, Samson Tan, Pedro~Ortiz Suarez, Victor Sanh, Hugo Lauren{\c{c}}on, Yacine Jernite, Julien Launay, Margaret Mitchell, Colin Raffel, Aaron Gokaslan, Adi Simhi, Aitor Soroa, Alham~Fikri Aji, Amit Alfassy, Anna Rogers, Ariel~Kreisberg Nitzav, Canwen Xu, Chenghao Mou, Chris Emezue, Christopher Klamm, Colin Leong, Daniel van Strien, David~Ifeoluwa Adelani, and et~al.
\newblock {BLOOM:} {A} 176b-parameter open-access multilingual language model.
\newblock \emph{CoRR}, abs/2211.05100, 2022.
\newblock \doi{10.48550/ARXIV.2211.05100}.
\newblock URL \url{https://doi.org/10.48550/arXiv.2211.05100}.

\bibitem[Touvron et~al.(2023{\natexlab{a}})Touvron, Lavril, Izacard, Martinet, Lachaux, Lacroix, Rozi{\`{e}}re, Goyal, Hambro, Azhar, Rodriguez, Joulin, Grave, and Lample]{DBLP:journals/corr/abs-2302-13971}
Hugo Touvron, Thibaut Lavril, Gautier Izacard, Xavier Martinet, Marie{-}Anne Lachaux, Timoth{\'{e}}e Lacroix, Baptiste Rozi{\`{e}}re, Naman Goyal, Eric Hambro, Faisal Azhar, Aur{\'{e}}lien Rodriguez, Armand Joulin, Edouard Grave, and Guillaume Lample.
\newblock Llama: Open and efficient foundation language models.
\newblock \emph{CoRR}, abs/2302.13971, 2023{\natexlab{a}}.
\newblock \doi{10.48550/ARXIV.2302.13971}.
\newblock URL \url{https://doi.org/10.48550/arXiv.2302.13971}.

\bibitem[Anthropic(2023{\natexlab{a}})]{claude}
Anthropic.
\newblock Introducing {Claude}, 2023{\natexlab{a}}.
\newblock URL \url{https://www.anthropic.com/index/introducing-claude}.

\bibitem[OpenAI(2023)]{DBLP:journals/corr/abs-2303-08774}
OpenAI.
\newblock {GPT-4} technical report.
\newblock \emph{CoRR}, abs/2303.08774, 2023.
\newblock \doi{10.48550/ARXIV.2303.08774}.
\newblock URL \url{https://doi.org/10.48550/arXiv.2303.08774}.

\bibitem[Anil et~al.(2023{\natexlab{a}})Anil, Dai, Firat, Johnson, Lepikhin, Passos, Shakeri, Taropa, Bailey, Chen, Chu, Clark, Shafey, Huang, Meier{-}Hellstern, Mishra, Moreira, Omernick, Robinson, Ruder, Tay, Xiao, Xu, Zhang, {\'{A}}brego, Ahn, Austin, Barham, Botha, Bradbury, Brahma, Brooks, Catasta, Cheng, Cherry, Choquette{-}Choo, Chowdhery, Crepy, Dave, Dehghani, Dev, Devlin, D{\'{\i}}az, Du, Dyer, Feinberg, Feng, Fienber, Freitag, Garcia, Gehrmann, Gonzalez, and et~al.]{DBLP:journals/corr/abs-2305-10403}
Rohan Anil, Andrew~M. Dai, Orhan Firat, Melvin Johnson, Dmitry Lepikhin, Alexandre Passos, Siamak Shakeri, Emanuel Taropa, Paige Bailey, Zhifeng Chen, Eric Chu, Jonathan~H. Clark, Laurent~El Shafey, Yanping Huang, Kathy Meier{-}Hellstern, Gaurav Mishra, Erica Moreira, Mark Omernick, Kevin Robinson, Sebastian Ruder, Yi~Tay, Kefan Xiao, Yuanzhong Xu, Yujing Zhang, Gustavo~Hern{\'{a}}ndez {\'{A}}brego, Junwhan Ahn, Jacob Austin, Paul Barham, Jan~A. Botha, James Bradbury, Siddhartha Brahma, Kevin Brooks, Michele Catasta, Yong Cheng, Colin Cherry, Christopher~A. Choquette{-}Choo, Aakanksha Chowdhery, Cl{\'{e}}ment Crepy, Shachi Dave, Mostafa Dehghani, Sunipa Dev, Jacob Devlin, Mark D{\'{\i}}az, Nan Du, Ethan Dyer, Vladimir Feinberg, Fangxiaoyu Feng, Vlad Fienber, Markus Freitag, Xavier Garcia, Sebastian Gehrmann, Lucas Gonzalez, and et~al.
\newblock Palm 2 technical report.
\newblock \emph{CoRR}, abs/2305.10403, 2023{\natexlab{a}}.
\newblock \doi{10.48550/ARXIV.2305.10403}.
\newblock URL \url{https://doi.org/10.48550/arXiv.2305.10403}.

\bibitem[Touvron et~al.(2023{\natexlab{b}})Touvron, Martin, Stone, Albert, Almahairi, Babaei, Bashlykov, Batra, Bhargava, Bhosale, Bikel, Blecher, Canton{-}Ferrer, Chen, Cucurull, Esiobu, Fernandes, Fu, Fu, Fuller, Gao, Goswami, Goyal, Hartshorn, Hosseini, Hou, Inan, Kardas, Kerkez, Khabsa, Kloumann, Korenev, Koura, Lachaux, Lavril, Lee, Liskovich, Lu, Mao, Martinet, Mihaylov, Mishra, Molybog, Nie, Poulton, Reizenstein, Rungta, Saladi, Schelten, Silva, Smith, Subramanian, Tan, Tang, Taylor, Williams, Kuan, Xu, Yan, Zarov, Zhang, Fan, Kambadur, Narang, Rodriguez, Stojnic, Edunov, and Scialom]{DBLP:journals/corr/abs-2307-09288}
Hugo Touvron, Louis Martin, Kevin Stone, Peter Albert, Amjad Almahairi, Yasmine Babaei, Nikolay Bashlykov, Soumya Batra, Prajjwal Bhargava, Shruti Bhosale, Dan Bikel, Lukas Blecher, Cristian Canton{-}Ferrer, Moya Chen, Guillem Cucurull, David Esiobu, Jude Fernandes, Jeremy Fu, Wenyin Fu, Brian Fuller, Cynthia Gao, Vedanuj Goswami, Naman Goyal, Anthony Hartshorn, Saghar Hosseini, Rui Hou, Hakan Inan, Marcin Kardas, Viktor Kerkez, Madian Khabsa, Isabel Kloumann, Artem Korenev, Punit~Singh Koura, Marie{-}Anne Lachaux, Thibaut Lavril, Jenya Lee, Diana Liskovich, Yinghai Lu, Yuning Mao, Xavier Martinet, Todor Mihaylov, Pushkar Mishra, Igor Molybog, Yixin Nie, Andrew Poulton, Jeremy Reizenstein, Rashi Rungta, Kalyan Saladi, Alan Schelten, Ruan Silva, Eric~Michael Smith, Ranjan Subramanian, Xiaoqing~Ellen Tan, Binh Tang, Ross Taylor, Adina Williams, Jian~Xiang Kuan, Puxin Xu, Zheng Yan, Iliyan Zarov, Yuchen Zhang, Angela Fan, Melanie Kambadur, Sharan Narang, Aur{\'{e}}lien Rodriguez, Robert Stojnic, Sergey Edunov,
  and Thomas Scialom.
\newblock Llama 2: Open foundation and fine-tuned chat models.
\newblock \emph{CoRR}, abs/2307.09288, 2023{\natexlab{b}}.
\newblock \doi{10.48550/ARXIV.2307.09288}.
\newblock URL \url{https://doi.org/10.48550/arXiv.2307.09288}.

\bibitem[Anthropic(2023{\natexlab{b}})]{claude2}
Anthropic.
\newblock Claude 2.
\newblock Technical report, Anthropic, 2023{\natexlab{b}}.
\newblock URL \url{https://www-files.anthropic.com/production/images/Model-Card-Claude-2.pdf}.

\bibitem[Bai et~al.(2023)Bai, Bai, Chu, Cui, Dang, Deng, Fan, Ge, Han, Huang, Hui, Ji, Li, Lin, Lin, Liu, Liu, Lu, Lu, Ma, Men, Ren, Ren, Tan, Tan, Tu, Wang, Wang, Wang, Wu, Xu, Xu, Yang, Yang, Yang, Yang, Yao, Yu, Yuan, Yuan, Zhang, Zhang, Zhang, Zhang, Zhou, Zhou, Zhou, and Zhu]{DBLP:journals/corr/abs-2309-16609}
Jinze Bai, Shuai Bai, Yunfei Chu, Zeyu Cui, Kai Dang, Xiaodong Deng, Yang Fan, Wenbin Ge, Yu~Han, Fei Huang, Binyuan Hui, Luo Ji, Mei Li, Junyang Lin, Runji Lin, Dayiheng Liu, Gao Liu, Chengqiang Lu, Keming Lu, Jianxin Ma, Rui Men, Xingzhang Ren, Xuancheng Ren, Chuanqi Tan, Sinan Tan, Jianhong Tu, Peng Wang, Shijie Wang, Wei Wang, Shengguang Wu, Benfeng Xu, Jin Xu, An~Yang, Hao Yang, Jian Yang, Shusheng Yang, Yang Yao, Bowen Yu, Hongyi Yuan, Zheng Yuan, Jianwei Zhang, Xingxuan Zhang, Yichang Zhang, Zhenru Zhang, Chang Zhou, Jingren Zhou, Xiaohuan Zhou, and Tianhang Zhu.
\newblock Qwen technical report.
\newblock \emph{CoRR}, abs/2309.16609, 2023.
\newblock \doi{10.48550/ARXIV.2309.16609}.
\newblock URL \url{https://doi.org/10.48550/arXiv.2309.16609}.

\bibitem[Jiang et~al.(2023)Jiang, Sablayrolles, Mensch, Bamford, Chaplot, de~Las~Casas, Bressand, Lengyel, Lample, Saulnier, Lavaud, Lachaux, Stock, Scao, Lavril, Wang, Lacroix, and Sayed]{DBLP:journals/corr/abs-2310-06825}
Albert~Q. Jiang, Alexandre Sablayrolles, Arthur Mensch, Chris Bamford, Devendra~Singh Chaplot, Diego de~Las~Casas, Florian Bressand, Gianna Lengyel, Guillaume Lample, Lucile Saulnier, L{\'{e}}lio~Renard Lavaud, Marie{-}Anne Lachaux, Pierre Stock, Teven~Le Scao, Thibaut Lavril, Thomas Wang, Timoth{\'{e}}e Lacroix, and William~El Sayed.
\newblock Mistral 7b.
\newblock \emph{CoRR}, abs/2310.06825, 2023.
\newblock \doi{10.48550/ARXIV.2310.06825}.
\newblock URL \url{https://doi.org/10.48550/arXiv.2310.06825}.

\bibitem[Almazrouei et~al.(2023)Almazrouei, Alobeidli, Alshamsi, Cappelli, Cojocaru, Debbah, Goffinet, Hesslow, Launay, Malartic, Mazzotta, Noune, Pannier, and Penedo]{DBLP:journals/corr/abs-2311-16867}
Ebtesam Almazrouei, Hamza Alobeidli, Abdulaziz Alshamsi, Alessandro Cappelli, Ruxandra Cojocaru, M{\'{e}}rouane Debbah, {\'{E}}tienne Goffinet, Daniel Hesslow, Julien Launay, Quentin Malartic, Daniele Mazzotta, Badreddine Noune, Baptiste Pannier, and Guilherme Penedo.
\newblock The falcon series of open language models.
\newblock \emph{CoRR}, abs/2311.16867, 2023.
\newblock \doi{10.48550/ARXIV.2311.16867}.
\newblock URL \url{https://doi.org/10.48550/arXiv.2311.16867}.

\bibitem[Anil et~al.(2023{\natexlab{b}})Anil, Borgeaud, Wu, Alayrac, Yu, Soricut, Schalkwyk, Dai, Hauth, Millican, Silver, Petrov, Johnson, Antonoglou, Schrittwieser, Glaese, Chen, Pitler, Lillicrap, Lazaridou, Firat, Molloy, Isard, Barham, Hennigan, Lee, Viola, Reynolds, Xu, Doherty, Collins, Meyer, Rutherford, Moreira, Ayoub, Goel, Tucker, Piqueras, Krikun, Barr, Savinov, Danihelka, Roelofs, White, Andreassen, von Glehn, Yagati, Kazemi, Gonzalez, Khalman, Sygnowski, and et~al.]{DBLP:journals/corr/abs-2312-11805}
Rohan Anil, Sebastian Borgeaud, Yonghui Wu, Jean{-}Baptiste Alayrac, Jiahui Yu, Radu Soricut, Johan Schalkwyk, Andrew~M. Dai, Anja Hauth, Katie Millican, David Silver, Slav Petrov, Melvin Johnson, Ioannis Antonoglou, Julian Schrittwieser, Amelia Glaese, Jilin Chen, Emily Pitler, Timothy~P. Lillicrap, Angeliki Lazaridou, Orhan Firat, James Molloy, Michael Isard, Paul~Ronald Barham, Tom Hennigan, Benjamin Lee, Fabio Viola, Malcolm Reynolds, Yuanzhong Xu, Ryan Doherty, Eli Collins, Clemens Meyer, Eliza Rutherford, Erica Moreira, Kareem Ayoub, Megha Goel, George Tucker, Enrique Piqueras, Maxim Krikun, Iain Barr, Nikolay Savinov, Ivo Danihelka, Becca Roelofs, Ana{\"{\i}}s White, Anders Andreassen, Tamara von Glehn, Lakshman Yagati, Mehran Kazemi, Lucas Gonzalez, Misha Khalman, Jakub Sygnowski, and et~al.
\newblock Gemini: {A} family of highly capable multimodal models.
\newblock \emph{CoRR}, abs/2312.11805, 2023{\natexlab{b}}.
\newblock \doi{10.48550/ARXIV.2312.11805}.
\newblock URL \url{https://doi.org/10.48550/arXiv.2312.11805}.

\bibitem[DeepSeek{-}AI et~al.(2024)DeepSeek{-}AI, Liu, Feng, Xue, Wang, Wu, Lu, Zhao, Deng, Zhang, Ruan, Dai, Guo, Yang, Chen, Ji, Li, Lin, Dai, Luo, Hao, Chen, Li, Zhang, Bao, Xu, Wang, Zhang, Ding, Xin, Gao, Li, Qu, Cai, Liang, Guo, Ni, Li, Wang, Chen, Chen, Yuan, Qiu, Li, Song, Dong, Hu, Gao, Guan, Huang, Yu, Wang, Zhang, Xu, Xia, Zhao, Wang, Zhang, Li, Wang, Zhang, Zhang, Tang, Li, Tian, Huang, Wang, Zhang, Wang, Zhu, Chen, Du, Chen, Jin, Ge, Zhang, Pan, Wang, Xu, Zhang, Chen, Li, Lu, Zhou, Chen, Wu, Ye, Ma, Wang, Zhou, Yu, Zhou, Pan, Wang, Yun, Pei, Sun, Xiao, and Zeng]{DBLP:journals/corr/abs-2412-19437}
DeepSeek{-}AI, Aixin Liu, Bei Feng, Bing Xue, Bingxuan Wang, Bochao Wu, Chengda Lu, Chenggang Zhao, Chengqi Deng, Chenyu Zhang, Chong Ruan, Damai Dai, Daya Guo, Dejian Yang, Deli Chen, Dongjie Ji, Erhang Li, Fangyun Lin, Fucong Dai, Fuli Luo, Guangbo Hao, Guanting Chen, Guowei Li, H.~Zhang, Han Bao, Hanwei Xu, Haocheng Wang, Haowei Zhang, Honghui Ding, Huajian Xin, Huazuo Gao, Hui Li, Hui Qu, J.~L. Cai, Jian Liang, Jianzhong Guo, Jiaqi Ni, Jiashi Li, Jiawei Wang, Jin Chen, Jingchang Chen, Jingyang Yuan, Junjie Qiu, Junlong Li, Junxiao Song, Kai Dong, Kai Hu, Kaige Gao, Kang Guan, Kexin Huang, Kuai Yu, Lean Wang, Lecong Zhang, Lei Xu, Leyi Xia, Liang Zhao, Litong Wang, Liyue Zhang, Meng Li, Miaojun Wang, Mingchuan Zhang, Minghua Zhang, Minghui Tang, Mingming Li, Ning Tian, Panpan Huang, Peiyi Wang, Peng Zhang, Qiancheng Wang, Qihao Zhu, Qinyu Chen, Qiushi Du, R.~J. Chen, R.~L. Jin, Ruiqi Ge, Ruisong Zhang, Ruizhe Pan, Runji Wang, Runxin Xu, Ruoyu Zhang, Ruyi Chen, S.~S. Li, Shanghao Lu, Shangyan Zhou,
  Shanhuang Chen, Shaoqing Wu, Shengfeng Ye, Shirong Ma, Shiyu Wang, Shuang Zhou, Shuiping Yu, Shunfeng Zhou, Shuting Pan, T.~Wang, Tao Yun, Tian Pei, Tianyu Sun, W.~L. Xiao, and Wangding Zeng.
\newblock Deepseek-v3 technical report.
\newblock \emph{CoRR}, abs/2412.19437, 2024.
\newblock \doi{10.48550/ARXIV.2412.19437}.
\newblock URL \url{https://doi.org/10.48550/arXiv.2412.19437}.

\bibitem[DeepSeek{-}AI et~al.(2025)DeepSeek{-}AI, Guo, Yang, Zhang, Song, Zhang, Xu, Zhu, Ma, Wang, Bi, Zhang, Yu, Wu, Wu, Gou, Shao, Li, Gao, Liu, Xue, Wang, Wu, Feng, Lu, Zhao, Deng, Zhang, Ruan, Dai, Chen, Ji, Li, Lin, Dai, Luo, Hao, Chen, Li, Zhang, Bao, Xu, Wang, Ding, Xin, Gao, Qu, Li, Guo, Li, Wang, Chen, Yuan, Qiu, Li, Cai, Ni, Liang, Chen, Dong, Hu, Gao, Guan, Huang, Yu, Wang, Zhang, Zhao, Wang, Zhang, Xu, Xia, Zhang, Zhang, Tang, Li, Wang, Li, Tian, Huang, Zhang, Wang, Chen, Du, Ge, Zhang, Pan, Wang, Chen, Jin, Chen, Lu, Zhou, Chen, Ye, Wang, Yu, Zhou, Pan, and Li]{DBLP:journals/corr/abs-2501-12948}
DeepSeek{-}AI, Daya Guo, Dejian Yang, Haowei Zhang, Junxiao Song, Ruoyu Zhang, Runxin Xu, Qihao Zhu, Shirong Ma, Peiyi Wang, Xiao Bi, Xiaokang Zhang, Xingkai Yu, Yu~Wu, Z.~F. Wu, Zhibin Gou, Zhihong Shao, Zhuoshu Li, Ziyi Gao, Aixin Liu, Bing Xue, Bingxuan Wang, Bochao Wu, Bei Feng, Chengda Lu, Chenggang Zhao, Chengqi Deng, Chenyu Zhang, Chong Ruan, Damai Dai, Deli Chen, Dongjie Ji, Erhang Li, Fangyun Lin, Fucong Dai, Fuli Luo, Guangbo Hao, Guanting Chen, Guowei Li, H.~Zhang, Han Bao, Hanwei Xu, Haocheng Wang, Honghui Ding, Huajian Xin, Huazuo Gao, Hui Qu, Hui Li, Jianzhong Guo, Jiashi Li, Jiawei Wang, Jingchang Chen, Jingyang Yuan, Junjie Qiu, Junlong Li, J.~L. Cai, Jiaqi Ni, Jian Liang, Jin Chen, Kai Dong, Kai Hu, Kaige Gao, Kang Guan, Kexin Huang, Kuai Yu, Lean Wang, Lecong Zhang, Liang Zhao, Litong Wang, Liyue Zhang, Lei Xu, Leyi Xia, Mingchuan Zhang, Minghua Zhang, Minghui Tang, Meng Li, Miaojun Wang, Mingming Li, Ning Tian, Panpan Huang, Peng Zhang, Qiancheng Wang, Qinyu Chen, Qiushi Du, Ruiqi Ge,
  Ruisong Zhang, Ruizhe Pan, Runji Wang, R.~J. Chen, R.~L. Jin, Ruyi Chen, Shanghao Lu, Shangyan Zhou, Shanhuang Chen, Shengfeng Ye, Shiyu Wang, Shuiping Yu, Shunfeng Zhou, Shuting Pan, and S.~S. Li.
\newblock Deepseek-r1: Incentivizing reasoning capability in llms via reinforcement learning.
\newblock \emph{CoRR}, abs/2501.12948, 2025.
\newblock \doi{10.48550/ARXIV.2501.12948}.
\newblock URL \url{https://doi.org/10.48550/arXiv.2501.12948}.

\bibitem[Anthropic(2025{\natexlab{a}})]{claude3.7}
Anthropic.
\newblock Claude 3.7 {Sonnet}, 2025{\natexlab{a}}.
\newblock URL \url{https://www.anthropic.com/news/claude-3-7-sonnet}.

\bibitem[OpenAI(2025)]{o3}
OpenAI.
\newblock {Introducing OpenAI o3 and o4-mini}, 2025.
\newblock URL \url{https://openai.com/index/introducing-o3-and-o4-mini/}.

\bibitem[{Meta AI}(2025)]{meta2025llama4}
{Meta AI}.
\newblock {The Llama 4 herd: The beginning of a new era of natively multimodal AI innovation}.
\newblock Blog, 2025.
\newblock URL \url{https://ai.meta.com/blog/llama-4-multimodal-intelligence/}.

\bibitem[xAI(2025)]{grok3}
xAI.
\newblock Grok 3 beta — the age of reasoning agents, 2025.
\newblock URL \url{https://x.ai/news/grok-3}.

\bibitem[Anthropic(2025{\natexlab{b}})]{claude4.5}
Anthropic.
\newblock Introducing claude sonnet 4.5, 2025{\natexlab{b}}.
\newblock URL \url{https://www.anthropic.com/news/claude-sonnet-4-5}.

\bibitem[Luccioni et~al.(2023)Luccioni, Viguier, and Ligozat]{DBLP:journals/jmlr/LuccioniVL23}
Alexandra~Sasha Luccioni, Sylvain Viguier, and Anne{-}Laure Ligozat.
\newblock Estimating the carbon footprint of bloom, a 176b parameter language model.
\newblock \emph{J. Mach. Learn. Res.}, 24:\penalty0 253:1--253:15, 2023.
\newblock URL \url{https://jmlr.org/papers/v24/23-0069.html}.

\bibitem[{NVIDIA} and {Meta}(2024)]{nvidia2024Llama31modelcard}
{NVIDIA} and {Meta}.
\newblock {Llama} 3.1 405b instruct model card.
\newblock Model documentation, NVIDIA Corporation, 2024.
\newblock URL \url{https://build.nvidia.com/meta/llama-3_1-405b-instruct/modelcard}.

\bibitem[Patterson et~al.(2021)Patterson, Gonzalez, Le, Liang, Munguia, Rothchild, So, Texier, and Dean]{DBLP:journals/corr/abs-2104-10350}
David~A. Patterson, Joseph Gonzalez, Quoc~V. Le, Chen Liang, Lluis{-}Miquel Munguia, Daniel Rothchild, David~R. So, Maud Texier, and Jeff Dean.
\newblock {Carbon Emissions and Large Neural Network Training}.
\newblock \emph{CoRR}, abs/2104.10350, 2021.
\newblock URL \url{https://arxiv.org/abs/2104.10350}.

\bibitem[Vaswani et~al.(2017)Vaswani, Shazeer, Parmar, Uszkoreit, Jones, Gomez, Kaiser, and Polosukhin]{DBLP:conf/nips/VaswaniSPUJGKP17}
Ashish Vaswani, Noam Shazeer, Niki Parmar, Jakob Uszkoreit, Llion Jones, Aidan~N. Gomez, Lukasz Kaiser, and Illia Polosukhin.
\newblock {Attention is All you Need}.
\newblock In Isabelle Guyon, Ulrike von Luxburg, Samy Bengio, Hanna~M. Wallach, Rob Fergus, S.~V.~N. Vishwanathan, and Roman Garnett, editors, \emph{Advances in Neural Information Processing Systems 30: Annual Conference on Neural Information Processing Systems 2017, December 4-9, 2017, Long Beach, CA, {USA}}, pages 5998--6008, 2017.
\newblock URL \url{https://proceedings.neurips.cc/paper/2017/hash/3f5ee243547dee91fbd053c1c4a845aa-Abstract.html}.

\bibitem[Dao et~al.(2022)Dao, Fu, Ermon, Rudra, and R{\'{e}}]{DBLP:conf/nips/DaoFERR22}
Tri Dao, Daniel~Y. Fu, Stefano Ermon, Atri Rudra, and Christopher R{\'{e}}.
\newblock {FlashAttention: Fast and Memory-Efficient Exact Attention with IO-Awareness}.
\newblock In Sanmi Koyejo, S.~Mohamed, A.~Agarwal, Danielle Belgrave, K.~Cho, and A.~Oh, editors, \emph{Advances in Neural Information Processing Systems 35: Annual Conference on Neural Information Processing Systems 2022, NeurIPS 2022, New Orleans, LA, USA, November 28 - December 9, 2022}, 2022.
\newblock URL \url{http://papers.nips.cc/paper\_files/paper/2022/hash/67d57c32e20fd0a7a302cb81d36e40d5-Abstract-Conference.html}.

\bibitem[{NVIDIA}(2023)]{nvidia2023tensorcore}
{NVIDIA}.
\newblock Tips for optimizing {GPU} performance using {Tensor Cores}.
\newblock NVIDIA Technical Blog, 2023.
\newblock URL \url{https://developer.nvidia.com/blog/optimizing-gpu-performance-tensor-cores/}.
\newblock Accessed: 2025-09-25.

\bibitem[Dao(2024)]{DBLP:conf/iclr/Dao24}
Tri Dao.
\newblock {FlashAttention-2: Faster Attention with Better Parallelism and Work Partitioning}.
\newblock In \emph{The Twelfth International Conference on Learning Representations, {ICLR} 2024, Vienna, Austria, May 7-11, 2024}. OpenReview.net, 2024.
\newblock URL \url{https://openreview.net/forum?id=mZn2Xyh9Ec}.

\bibitem[Shah et~al.(2024)Shah, Bikshandi, Zhang, Thakkar, Ramani, and Dao]{DBLP:conf/nips/ShahBZTRD24}
Jay Shah, Ganesh Bikshandi, Ying Zhang, Vijay Thakkar, Pradeep Ramani, and Tri Dao.
\newblock {FlashAttention-3: Fast and Accurate Attention with Asynchrony and Low-precision}.
\newblock In Amir Globersons, Lester Mackey, Danielle Belgrave, Angela Fan, Ulrich Paquet, Jakub~M. Tomczak, and Cheng Zhang, editors, \emph{Advances in Neural Information Processing Systems 38: Annual Conference on Neural Information Processing Systems 2024, NeurIPS 2024, Vancouver, BC, Canada, December 10 - 15, 2024}, 2024.
\newblock URL \url{http://papers.nips.cc/paper\_files/paper/2024/hash/7ede97c3e082c6df10a8d6103a2eebd2-Abstract-Conference.html}.

\bibitem[Fedus et~al.(2022)Fedus, Zoph, and Shazeer]{DBLP:journals/jmlr/FedusZS22}
William Fedus, Barret Zoph, and Noam Shazeer.
\newblock {Switch Transformers: Scaling to Trillion Parameter Models with Simple and Efficient Sparsity}.
\newblock \emph{J. Mach. Learn. Res.}, 23:\penalty0 120:1--120:39, 2022.
\newblock URL \url{https://jmlr.org/papers/v23/21-0998.html}.

\bibitem[{NVIDIA Corporation}(2024)]{nvidia2024h200}
{NVIDIA Corporation}.
\newblock \emph{{NVIDIA H200 Tensor Core GPU} Specifications}, 2024.
\newblock URL \url{https://www.nvidia.com/en-us/data-center/h200/}.

\bibitem[Soboleva et~al.(2023)Soboleva, Al-Khateeb, Myers, Steeves, Hestness, and Dey]{cerebras2023slimpajama}
Daria Soboleva, Faisal Al-Khateeb, Robert Myers, Jacob~R Steeves, Joel Hestness, and Nolan Dey.
\newblock {SlimPajama: A 627B token cleaned and deduplicated version of RedPajama}.
\newblock \url{https://cerebras.ai/blog/slimpajama-a-627b-token-cleaned-and-deduplicated-version-of-redpajama}, 2023.
\newblock URL \url{https://huggingface.co/datasets/cerebras/SlimPajama-627B}.

\bibitem[Weber et~al.(2024)Weber, Fu, Anthony, Oren, Adams, Alexandrov, Lyu, Nguyen, Yao, Adams, Athiwaratkun, Chalamala, Chen, Ryabinin, Dao, Liang, R{\'{e}}, Rish, and Zhang]{DBLP:conf/nips/WeberFAOAALNYAA24}
Maurice Weber, Daniel~Y. Fu, Quentin Anthony, Yonatan Oren, Shane Adams, Anton Alexandrov, Xiaozhong Lyu, Huu Nguyen, Xiaozhe Yao, Virginia Adams, Ben Athiwaratkun, Rahul Chalamala, Kezhen Chen, Max Ryabinin, Tri Dao, Percy Liang, Christopher R{\'{e}}, Irina Rish, and Ce~Zhang.
\newblock {RedPajama: an Open Dataset for Training Large Language Models}.
\newblock In Amir Globersons, Lester Mackey, Danielle Belgrave, Angela Fan, Ulrich Paquet, Jakub~M. Tomczak, and Cheng Zhang, editors, \emph{Advances in Neural Information Processing Systems 38: Annual Conference on Neural Information Processing Systems 2024, NeurIPS 2024, Vancouver, BC, Canada, December 10 - 15, 2024}, 2024.
\newblock URL \url{http://papers.nips.cc/paper\_files/paper/2024/hash/d34497330b1fd6530f7afd86d0df9f76-Abstract-Datasets\_and\_Benchmarks\_Track.html}.

\bibitem[Kudo and Richardson(2018)]{DBLP:conf/emnlp/KudoR18}
Taku Kudo and John Richardson.
\newblock Sentencepiece: {A} simple and language independent subword tokenizer and detokenizer for neural text processing.
\newblock In Eduardo Blanco and Wei Lu, editors, \emph{Proceedings of the 2018 Conference on Empirical Methods in Natural Language Processing, {EMNLP} 2018: System Demonstrations, Brussels, Belgium, October 31 - November 4, 2018}, pages 66--71. Association for Computational Linguistics, 2018.
\newblock \doi{10.18653/V1/D18-2012}.
\newblock URL \url{https://doi.org/10.18653/v1/d18-2012}.

\bibitem[Gage(1994)]{10.5555/177910.177914}
Philip Gage.
\newblock A new algorithm for data compression.
\newblock \emph{C Users J.}, 12\penalty0 (2):\penalty0 23–38, February 1994.
\newblock ISSN 0898-9788.

\bibitem[Zhang and Sennrich(2019)]{DBLP:conf/nips/ZhangS19a}
Biao Zhang and Rico Sennrich.
\newblock Root mean square layer normalization.
\newblock In Hanna~M. Wallach, Hugo Larochelle, Alina Beygelzimer, Florence d'Alch{\'{e}}{-}Buc, Emily~B. Fox, and Roman Garnett, editors, \emph{Advances in Neural Information Processing Systems 32: Annual Conference on Neural Information Processing Systems 2019, NeurIPS 2019, December 8-14, 2019, Vancouver, BC, Canada}, pages 12360--12371, 2019.
\newblock URL \url{https://proceedings.neurips.cc/paper/2019/hash/1e8a19426224ca89e83cef47f1e7f53b-Abstract.html}.

\bibitem[Shazeer(2020)]{DBLP:journals/corr/abs-2002-05202}
Noam Shazeer.
\newblock {GLU} variants improve transformer.
\newblock \emph{CoRR}, abs/2002.05202, 2020.
\newblock URL \url{https://arxiv.org/abs/2002.05202}.

\bibitem[Su et~al.(2024)Su, Ahmed, Lu, Pan, Bo, and Liu]{DBLP:journals/ijon/SuALPBL24}
Jianlin Su, Murtadha H.~M. Ahmed, Yu~Lu, Shengfeng Pan, Wen Bo, and Yunfeng Liu.
\newblock Roformer: Enhanced transformer with rotary position embedding.
\newblock \emph{Neurocomputing}, 568:\penalty0 127063, 2024.
\newblock \doi{10.1016/J.NEUCOM.2023.127063}.
\newblock URL \url{https://doi.org/10.1016/j.neucom.2023.127063}.

\bibitem[Ainslie et~al.(2023)Ainslie, Lee{-}Thorp, de~Jong, Zemlyanskiy, Lebr{\'{o}}n, and Sanghai]{DBLP:conf/emnlp/AinslieLJZLS23}
Joshua Ainslie, James Lee{-}Thorp, Michiel de~Jong, Yury Zemlyanskiy, Federico Lebr{\'{o}}n, and Sumit Sanghai.
\newblock {GQA:} training generalized multi-query transformer models from multi-head checkpoints.
\newblock In Houda Bouamor, Juan Pino, and Kalika Bali, editors, \emph{Proceedings of the 2023 Conference on Empirical Methods in Natural Language Processing, {EMNLP} 2023, Singapore, December 6-10, 2023}, pages 4895--4901. Association for Computational Linguistics, 2023.
\newblock \doi{10.18653/V1/2023.EMNLP-MAIN.298}.
\newblock URL \url{https://doi.org/10.18653/v1/2023.emnlp-main.298}.

\bibitem[Tazi et~al.(2025)Tazi, Mom, Zhao, Nguyen, Mekkouri, Werra, and Wolf]{ultrascale_playbook}
Nouamane Tazi, Ferdinand Mom, Haojun Zhao, Phuc Nguyen, Mohamed Mekkouri, Leandro Werra, and Thomas Wolf.
\newblock {The Ultra-Scale Playbook: Training LLMs on GPU Clusters}, 2025.

\bibitem[Liang et~al.(2025)Liang, Liu, Wright, Constable, Gu, Huang, Zhang, Feng, Huang, Wang, Purandare, Nadathur, and Idreos]{liang2025torchtitan}
Wanchao Liang, Tianyu Liu, Less Wright, Will Constable, Andrew Gu, Chien-Chin Huang, Iris Zhang, Wei Feng, Howard Huang, Junjie Wang, Sanket Purandare, Gokul Nadathur, and Stratos Idreos.
\newblock {TorchTitan: One-stop PyTorch native solution for production ready {LLM} pretraining}.
\newblock In \emph{The Thirteenth International Conference on Learning Representations}, 2025.
\newblock URL \url{https://openreview.net/forum?id=SFN6Wm7YBI}.

\bibitem[Loshchilov and Hutter(2019)]{DBLP:conf/iclr/LoshchilovH19}
Ilya Loshchilov and Frank Hutter.
\newblock Decoupled weight decay regularization.
\newblock In \emph{7th International Conference on Learning Representations, {ICLR} 2019, New Orleans, LA, USA, May 6-9, 2019}. OpenReview.net, 2019.
\newblock URL \url{https://openreview.net/forum?id=Bkg6RiCqY7}.

\bibitem[Rajbhandari et~al.(2020)Rajbhandari, Rasley, Ruwase, and He]{DBLP:conf/sc/RajbhandariRRH20}
Samyam Rajbhandari, Jeff Rasley, Olatunji Ruwase, and Yuxiong He.
\newblock Zero: memory optimizations toward training trillion parameter models.
\newblock In Christine Cuicchi, Irene Qualters, and William~T. Kramer, editors, \emph{Proceedings of the International Conference for High Performance Computing, Networking, Storage and Analysis, {SC} 2020, Virtual Event / Atlanta, Georgia, USA, November 9-19, 2020}, page~20. {IEEE/ACM}, 2020.
\newblock \doi{10.1109/SC41405.2020.00024}.
\newblock URL \url{https://doi.org/10.1109/SC41405.2020.00024}.

\bibitem[Biewald(2020)]{wandb}
Lukas Biewald.
\newblock Experiment tracking with weights and biases, 2020.
\newblock URL \url{https://www.wandb.ai/}.
\newblock Software available from wandb.ai.

\bibitem[{U.S. Environmental Protection Agency}(2024)]{epa_ghg_calc_2024}
{U.S. Environmental Protection Agency}.
\newblock Greenhouse gas equivalencies calculator - calculations and references.
\newblock \url{https://www.epa.gov/energy/greenhouse-gas-equivalencies-calculator-calculations-and-references}, 2024.
\newblock Accessed: September 25, 2025.

\bibitem[{U.S. Environmental Protection Agency}(2025{\natexlab{a}})]{epa_egrid_2025}
{U.S. Environmental Protection Agency}.
\newblock Emissions \& generation resource integrated database (egrid) summary data, 2023 data.
\newblock \url{https://www.epa.gov/egrid/summary-data}, January 2025{\natexlab{a}}.
\newblock Released: January 15, 2025.

\bibitem[{U.S. Environmental Protection Agency}(2025{\natexlab{b}})]{epa_vehicle_emissions_2025}
{U.S. Environmental Protection Agency}.
\newblock Greenhouse gas emissions from a typical passenger vehicle.
\newblock \url{https://www.epa.gov/greenvehicles/greenhouse-gas-emissions-typical-passenger-vehicle}, 2025{\natexlab{b}}.
\newblock Accessed: September 25, 2025.

\bibitem[Rafailov et~al.(2023)Rafailov, Sharma, Mitchell, Manning, Ermon, and Finn]{DBLP:conf/nips/RafailovSMMEF23}
Rafael Rafailov, Archit Sharma, Eric Mitchell, Christopher~D. Manning, Stefano Ermon, and Chelsea Finn.
\newblock Direct preference optimization: Your language model is secretly a reward model.
\newblock In Alice Oh, Tristan Naumann, Amir Globerson, Kate Saenko, Moritz Hardt, and Sergey Levine, editors, \emph{Advances in Neural Information Processing Systems 36: Annual Conference on Neural Information Processing Systems 2023, NeurIPS 2023, New Orleans, LA, USA, December 10 - 16, 2023}, 2023.
\newblock URL \url{http://papers.nips.cc/paper\_files/paper/2023/hash/a85b405ed65c6477a4fe8302b5e06ce7-Abstract-Conference.html}.

\end{thebibliography}

\pagebreak

\end{document}